\documentclass[journal]{IEEEtran}
\usepackage{amsmath,graphicx} 
\usepackage{booktabs, multicol, multirow} % for excel2latex
\usepackage{color,soul} % highlighting 
\soulregister\ref{7} % make soul work with references
\soulregister\cite{7} % make soul work with citations

\usepackage[colorinlistoftodos]{todonotes} % MS-style ToDo notes
\usepackage{rotating} % sideways tables
\usepackage{tabularx} % better tables
\usepackage{float} % force figures in place
\usepackage{hyperref}

\usepackage{times}
\usepackage{epsfig}
\usepackage{graphicx}
\usepackage{amssymb}
\usepackage{stfloats}
\usepackage{setspace}

\def\Npatientsbonelesions{59}
\def\Npatientslymphnodes{176}
\def\Npatientspolyps{1,186}
\def\Npatientspolypstraining{394}
\def\Npatientspolypstesting{792}

\soulregister\Npatientsbonelesions{7}
\soulregister\Npatientslymphnodes{7}
\soulregister\Npatientspolyps{7}
\soulregister\Npatientspolypstraining{7}
\soulregister\Npatientspolypstesting{7}

\begin{document}
%
%\title{Improving computer-aided detection performance using random sets of deep convolutional network observations}
\title{Improving Computer-aided Detection using Convolutional Neural Networks and Random View Aggregation}
\author{Holger R. Roth, Le Lu, {\em Senior Member, IEEE}, Jiamin Liu, Jianhua Yao, Ari Seff, Kevin Cherry, Lauren Kim, and Ronald M. Summers% <-this % stops a space
\thanks{All authors are with the Imaging Biomarkers and Computer-Aided Diagnosis Laboratory, Radiology and Imaging Sciences Department, National Institutes of Health Clinical Center, Bethesda, MD 20892-1182, USA, e-mail: holger.roth@nih.gov.% <-this % stops a space
}} % \author
\maketitle
\begin{abstract}Automated computer-aided detection (CADe) in medical imaging has been an important tool in clinical practice and research. State-of-the-art methods often show high sensitivities but at the cost of high false-positives (FP)  per patient rates. We design a two-tiered coarse-to-fine cascade framework that first operates a candidate generation system at sensitivities of $\sim$100\% but at high FP levels. By leveraging existing CAD systems, coordinates of regions or volumes of interest (ROI or VOI) for lesion candidates are generated in this step and function as input for a second tier, which is our focus in this study. In this second stage, we generate $N$ 2D (two-dimensional) or 2.5D views via sampling through scale transformations, random translations and rotations with respect to each ROI's centroid coordinates. These random views are used to train deep convolutional neural network (ConvNet) classifiers. In testing, the trained ConvNets are employed to assign class (e.g., lesion, pathology) probabilities for a new set of $N$ random views that are then averaged at each ROI to compute a final per-candidate classification probability. This second tier behaves as a highly selective process to reject difficult false positives while preserving high sensitivities. The methods are evaluated on three different data sets with different numbers of patients: {\Npatientsbonelesions} patients for  sclerotic metastases detection, {\Npatientslymphnodes} patients for lymph node detection, and {\Npatientspolyps} patients for colonic polyp detection. Experimental results show the ability of ConvNets to generalize well to different medical imaging CADe applications and scale elegantly to various data sets. Our proposed methods improve CADe performance markedly in all cases. CADe sensitivities improved from 57\% to 70\%, from 43\% to 77\% and from 58\% to 75\% at 3 FPs per patient for sclerotic metastases, lymph nodes and colonic polyps, respectively.
\end{abstract}
%###################################################################################################
\IEEEpeerreviewmaketitle
\section{Introduction}
\IEEEPARstart{A}{ccurate} computer-aided detection (CADe) plays a central role in radiological diagnoses. The early detection of abnormal anatomies or precursors of pathology associated with cancer can aid in preventing the disease, which is among the leading causes of death worldwide \cite{who2014cancer}. Furthermore, detection can help to assess the staging of a patient's disease, and thus has the potential to alter a patient’s required treatment regimen \cite{msaouel2008mechanisms}. Computed tomography (CT), a ubiquitous screening and staging modality employed for disease detection in cancer patients, is commonly used for the detection of abnormal anatomy such as tumors and their metastases. At present, the detection of an abnormal anatomy via CT often occurs during manual prospective visual inspection of every image slice (of which there may be thousands) and every section of every image in each patient's CT study. This is a complex process that, when performed under a time restriction, is prone to error. Thorough manual assessment and processing is time-consuming and often delays the clinical workflow. Therefore CADe has the potential to greatly reduce the radiologists' clinical workload and to serve as a first or second reader for improved assessment of the disease \cite{wiese2012detection,burns2013automated,hammon2013automatic}.

CADe has been an active research area in medical imaging for the last two decades. Most work is based on some type of image feature extractor that is computed in a region-of-interest (ROI) in the image, e.g. intensity statistics, histogram of oriented gradients (HoG) \cite{seff20142d}, scale-invariant feature transform (SIFT) \cite{Toews2007statistical}, Hessian based shape descriptors (such as blobness) \cite{Wu2010stratified}, etc. These features are then used to learn a binary or discrete classifier, commonly linear support vector machines (SVM) and random forests, to differentiate normal from abnormal anatomy. At present, examples of CADe used in clinical practice include polyp detection for colon cancer screening \cite{summers2002colonic,Ravesteijn2010}, lung nodule detection for lung cancer screening \cite{Ginneken2015,firmino2014computer} or breast cancer screening with mammography \cite{cheng2003computer}. However, many applications of CADe result in significantly low sensitivity and/or specificity levels (i.e. high numbers of false negatives or false positives per volume). For this reason, they have not yet been incorporated into clinical practice. 

The method presented here aims to build upon existing CADe systems by forming a hierarchical two-tiered CADe system, designed to improve overall detection performance (i.e., high recalls together with low, or manageable FP rates per patient). To this end, we propose a new representation that efficiently integrates recent advances in computer vision, namely deep convolutional neural networks \cite{krizhevsky2012imagenet,lecun1989backpropagation} (ConvNets, see Fig. \ref{fig:convnet}).  

Recently, the availability of large amounts of annotated training sets and the accessibility of affordable parallel computing resources via Graphics Processing Units (or GPUs) have made it feasible to train deep convolutional neural networks (ConvNets). ConvNets have popularized the topic of ``deep learning'' in computer vision research \cite{jones2014computer}. The usage of ConvNets has allowed for substantial advancements not only in the classification of natural images \cite{krizhevsky2012imagenet}, but also in biomedical applications, such as mitosis detection in digital pathology \cite{cirecsan2013mitosis,ciresan2012deep}. Additionally, recent work has shown how the implementation of ConvNets can substantially improve the performance of state-of-the-art CADe systems \cite{prasoon2013deep, roth2014new, roth2014detection, li2014medical}. For instance, \cite{prasoon2013deep} proposes an MRI-based knee cartilage segmentation using a triplanar ConvNet. \cite{turaga2010convolutional} describes a supervised 3D boundary detection in volumetric electron microscopy (EM) images via ConvNets. 

In this study, we apply ConvNets along with random sets of 2D or 2.5D sampled views or observations. Our work partly draws upon the idea of hybrid systems, which use both parametric and non-parametric models for hierarchical coarse-to-fine classification. \cite{lu2011coarse}. The non-parametric model is replaced with aggregating decisions via ConvNets performed on random views.

Our contributions are the following: 

1) We propose a universal 2.5D image decomposition representation for utilizing ConvNets in CADe problems which can be generalized to others (with randomly sampled views or sampled under some problem-specific constraints, e.g., using local vessel orientations); 
2) we propose a new random aggregation method based on the deep ConvNet classification approach; 
3) we validate on three different datasets with different numbers of patients and CADe applications; 

and 4) markedly improve performance in all three cases. In particular, we improve CADe sensitivities from 57\% to 70\%, from 43\% to 77\% and 58\% to 75\% at 3 FPs per patient for sclerotic metastases \cite{burns2013automated}, lymph nodes \cite{cherry2014abdominal,liu2014mediastinal} and colonic polyps \cite{summers2005computed,Ravesteijn2010}, respectively. This  paper extends our preliminary work on lymph node \cite{roth2014new} and sclerotic bone metastasis detection \cite{roth2014detection} and includes performance evaluation on a new data set for detecting 252 colonic polyps in 1,186 patients. We show how ConvNets can be applied to build more accurate classifiers for CADe systems, as an effective false positive pruning process while maintaining high sensitivity recalls.
%################################################################################## 
\section{Methods}
Here, we describe our methods in detail. First, deep convolutional networks (ConvNets) are introduced, then we describe how to apply ConvNets to CADe application in a 2D or 2.5D approach and how to utilize random ConvNet observations in the fashion of a decompositional representation. Lastly, we describe various ways of candidate generation (CG) that are applicable for the using ConvNets on different data sets.
%##################################################################################
\subsection{Convolutional Neural Networks}
\label{sec:convnet}
\begin{figure}[!htb]
	\centering
		\includegraphics[width=8.5cm]{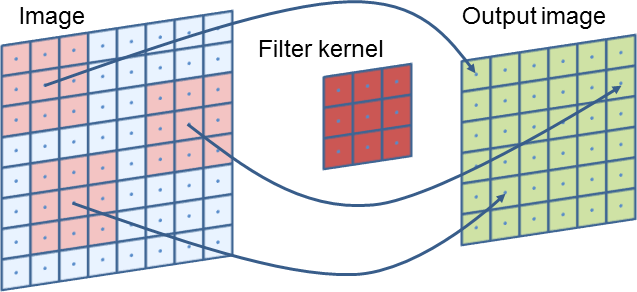}
	\caption{Features are computed by convolving filter kernels over the input region of interest. The input image can be padded to produce convolution responses of the same size as the input image.}
	\label{fig:convolution}
\end{figure}

\begin{figure}[!htb]
	\centering
		\includegraphics[width=4.5cm]{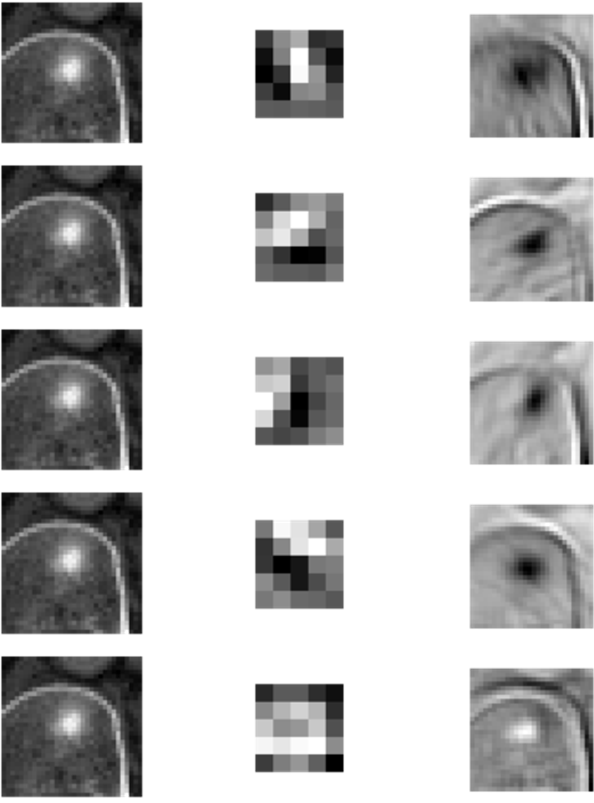}
	\caption{Some examples of filter responses (Right) after convolution with trained ConvNet kernels (Middle) of the first layer (showing an example of a sclerotic bone lesion in CT (Left).}
	\label{fig:conv1_responses}
\end{figure}

ConvNets are named for their convolutional filters that are used to compute image features for classification (see Fig. \ref{fig:convolution}). In this work, we use two cascaded layers of convolutional filters. All convolutional filter kernel elements are trained from the data in a supervised fashion by learning from a labeled set of examples. This has major advantages over more traditional CADe approaches that use hand-crafted features, designed from human experience. ConvNets have a better chance of capturing the ``essence'' of the imaging data set used for training than do hand-crafted features \cite{jones2014computer,seff20142d,Toews2007statistical,Wu2010stratified}. Furthermore, we can train similarly configured ConvNet architectures from randomly initialized or pre-trained model parameters for detecting different lesions or pathologies (with heterogeneous appearances), with no manual intervention of system and feature design. Examples of trained filters of the first convolutional layer and their responses are shown in Fig. \ref{fig:conv1_responses}.
%##################################################################################
\begin{figure}[!htb]
	\centering
		\includegraphics[width=6.5cm]{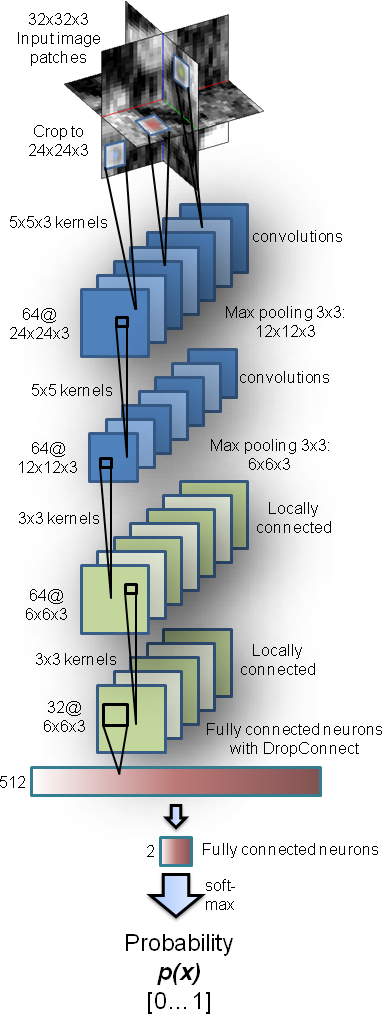}
	\caption{ConvNet applied to a 2.5D volume of interest extracted from a CT image. The number of convolutional filters, kernel sizes, and neural network connections for each layer are as shown. We use overlapping kernels with stride 2 during max-pooling.}
	\label{fig:convnet}
\end{figure}
%################################################################################## 
%################################################################################## 
%################################################################################## 
In-between convolutional layers, the ConvNet performs \textit{max-pooling} operations in order to summarize feature responses across neighboring pixels (see Fig. \ref{fig:convnet}). Such operations allow the ConvNet to learn features that are spatially invariant with respect to the location of objects in the images. Feature responses after the second convolutional layer feed into two \textit{locally connected} layers (similar to a convolutional layer but without weight sharing), and then \textit{fully-connected} neural network layers for classification. The deeper the convolutional layers in a ConvNet, the higher the order of image features they encode. This neural network learns how to interpret the feature responses and performs classifications. Our ConvNet uses a final \textit{softmax} layer which provides a classification probability for each input image (see Fig. \ref{fig:convnet}). In order to avoid overfitting, the fully-connected layers are constrained, using the \textit{``DropConnect''} method \cite{wan2013regularization}. \textit{DropConnect} behaves as a regularizer when training the ConvNet by preventing co-adaptation of units in the neural network. It is a variation of the previously suggested \textit{``DropOut''} method \cite{hinton2012improving,srivastava2014dropout}. We use and modify an open-source implementation (\textit{cuda-convnet}\footnote{\url{https://code.google.com/p/cuda-convnet}}) by Krizhevsky et al. \cite{krizhevsky2012imagenet,krizhevsky2014one} which efficiently trains the ConvNet by using GPU acceleration with the DropConnect modification by \cite{wan2013regularization}. Additional speed-ups are achieved by using rectified linear units as neuron activation functions, as opposed to the functions $f(x) = \tanh(x)$ or $f(x) = (1 + e^{-x})^{-1}$ from traditional neuron models, in the training and evaluation phases \cite{krizhevsky2012imagenet}. The input image can be cropped in order to train on translations of the cropped input image for data augmentation \cite{krizhevsky2012imagenet}. Our ConvNets are trained using stochastic gradient descent with momentum for 700-300-100-100 epochs on mini-batches of 64-64-32-16 images similar to \cite{wan2013regularization} on the CIFAR-10 data set (using an initial learning rate of 0.001 with the default weight decay). The per-pixel mean of the training image set is subtracted from each image fed to the ConvNet.
%################################################################################## 
\subsection{Applying ConvNets to CADe -- a 2D or 2.5D Approach}
\label{sec:roi_voi}
\begin{figure}[!htb]
	\centering
		\includegraphics[width=5.5cm]{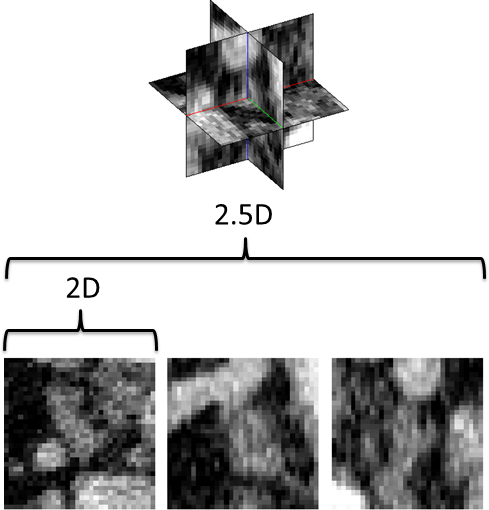}
	\caption{CADe locations can be either observed as 2D image patches or using a 2.5D approach, that samples the image using three orthogonal views. Here, a lymph node in CT is shown as the input to our method.}
	\label{fig:roi_voi}
\end{figure}

Depending on the imaging data, we explore a two-dimensional (2D) or two-and-a-half-dimensional (2.5D) representation to compute a ConvNet observation, sampled at each CADe candidate location (see Fig. \ref{fig:roi_voi}). In 2D, we refer to extracting a Region-of-Interest (ROI). In 2.5D, we refer to extracting a Volume-of-Interest (VOI). CADe candidate locations are normally obtained by a candidate generation process, which requires very high (i.e., close to 100\%) sensitivity at high false positives per patient or volume ($40\sim60$ FPs for our lymph node or bone lesion data sets and $\sim150$ FPs in colonic polyp cases). This performance standard can be easily attained by existing work \cite{burns2013automated,liu2014mediastinal,cherry2014abdominal,summers2005computed}.
%################################################################################## 
%################################################################################## 
\subsection{Random ConvNet Observations}
 In order to increase the variation of the training data and to avoid overfitting analogous to the data augmentation approach in \cite{krizhevsky2012imagenet,cirecsan2013mitosis} and \cite{ciresan2012deep}, multiple 2D or 2.5D observations per ROI or VOI are needed, respectively. Each ROI/VOI can be translated along a random vector $v$ in the CT space $N_t$ times. Furthermore, each translated ROI is rotated around its center $N_r$ times by a random angle $\alpha = [0^{\circ},\ldots,360^{\circ}]$. These translations and rotations for each ROI are computed $N_s$ times at different physical scales $s$ (the edge length of each ROI\footnote{Without loss of generality, the sampled 2D or 2.5D image patches or observations have the squared shape.}), but with fixed numbers of pixels by resampling (i.e., the physical pixel size will vary in the units of millimeters against different $s$). This procedure results in $N = N_s\times N_t\times N_r$ random observations of each ROI -- an approach similar to \cite{gokturk01astatistical}. Only 2D reformatting and sampling representation within an axial CT slice (axial reconstruction is the most common CT reconstruction imaging protocol) is employed when the inter-slice distances or slice thicknesses are 5mm or more. Following this procedure, both the training and test data sets can be expanded to larger scales, which will enhance the neural net’s generality and trainability. A ConvNet's predictions on these $N$ random observations $\left\{P_1(x),\ldots,P_N\right\}$ can then be simply averaged\footnote{We empirically evaluate several aggregation schemes on computing the final candidate class probability from a collection of ConvNet observations. Simple average performs the best and has good efficiency.} at each ROI to compute a per-candidate probability:
%################################################################################## 
\begin{equation}
	p\left(x|\{P_1(x),\ldots,P_N(x)\}\right) = \frac{1}{N}\sum_{i=1}^{N}P_i(x).
	\label{equ:prob}
\end{equation}
%################################################################################## 
Here, $P_i(x)$ is the ConvNet's classification probability computed for one individual 2D or 2.5D image patch. In theory, more sophisticated fusion rules can be explored, but simple averaging has proven to be effective for this experiment \cite{roth2014new}. 
%################################################################################## 
\begin{figure}[!htb]
	\centering
		\includegraphics[width=8.5cm]{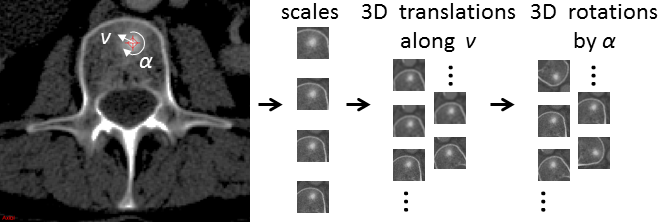}
	\caption{Image patches are generated from CADe candidates using different scales, 2D/3D translations (along a random vector $v$) and rotations (by a random angle $\alpha$) in the axial/3D plane (The example shows a sclerotic bone lesion in CT).}
	\label{fig:random_observations}
\end{figure}
%################################################################################## 
Furthermore, this random resampling method simply and effectively increases the amount of available training data. In computer vision, translational shifting and mirroring of 2D image patches are often used for this purpose \cite{krizhevsky2012imagenet}. By averaging the $N$ predictions on random 2D or 2.5D views as in Eq. \ref{equ:prob}, the robustness and stability of ConvNet can be further increased in testing, as shown in Sec. \ref{sec:results}. 
%################################################################################## 
%###################################################################################
\subsection{Candidate Generation}
\label{sec:candidates}
In general, any CADe system with a reasonably high sensitivity level (e.g., $\sim95\%$) at an acceptable FP rate (e.g., $\leq150$ per patient) can be used as a candidate location generation step in our proposed framework. Based on a reference data set, such a candidate can be then labeled as a `positive' or `negative' example and used to train a ConvNet. In this paper, we propose to apply the ConvNet as a second, more accurate classifier. This is a coarse-to-fine classification approach slightly inspired by other CADe schemes such as presented in \cite{lu2011coarse} although our methods are significantly different. 

In this study, we use three existing CADe systems that have previously been described in the literature:
\paragraph{Detection of sclerotic spine metastases} we use a recent CADe method for detecting sclerotic metastases candidates from  CT volumes \cite{burns2013automated,wiese2011computer} (see Sec. \ref{sec:results_bonelesions}). The spine is initially segmented by thresholding at certain CT attenuation levels and performing region growing. Furthermore, morphological operations are used to refine the segmentation and allow the extraction of the spinal canal. Further information on spine canal segmentation and partitioning is provided in \cite{yao2006automated}. Axial 2D cross sections of the vertebrae are then divided into sub-segments by a watershed algorithm based on local density differences \cite{yao2006computer}. The CADe algorithm then finds initial detections that have higher mean attenuation levels, in contrast to their neighboring 2D sub-segments. Since the watershed algorithm may over-segment the image, similar 2D sub-segment detections are merged by performing an energy minimization based on graph-cut and attenuation thresholds. Finally, 2D detections on neighboring cross sections are combined to form 3D detections with a graph-cut based merger. Each 3D detection acts as a seed point for a level-set segmentation method that segments the lesions in 3D. This step allows us to compute 25 characteristic features, such as shape, size, location, attenuation, volume, and sphericity. Finally, a committee of SVMs \cite{yao2005optimizing} is trained on these features. 

\paragraph{Detection of lymph nodes} we employ two preliminary CADe systems for detecting lymph node candidates from mediastinal \cite{liu2014mediastinal} and abdominal \cite{cherry2014abdominal} body regions (see Sec. \ref{sec:results_lymphnodes}), respectively. In the mediastinum, lungs are segmented automatically and shape features are computed at the voxel-level. The system uses a spatial prior of anatomical structures (such as the esophagus, aortic arch, and/or heart) via multi-atlas label fusion before detecting lymph node candidates using a SVM for classification. In the abdomen, a random forest classifier is used to create voxel-level lymph node predictions via image features. Both systems permit the combination of multiple statistical image descriptors (such as Hessian blobness and HOG) and appropriate feature selection in order to improve lymph node detection beyond traditional enhancement filters. Currently, 94\%-97\% sensitivity levels at rates of 25-35 FP/vol. can be achieved (\cite{liu2014mediastinal,cherry2014abdominal}). With sufficient training in the lymph node candidate generation step, close to 100\% sensitivities could be reached in the future.

\paragraph{Detection of colonic polyps} we apply a candidate generation step using the CADe system presented in \cite{summers2005computed} (see Sec. \ref{sec:resutls_polyps}). In this system, the colonic wall and lumen are first segmented, and any tagged colonic fluids are removed from CT colonography (CTC) volumes. In order to identify colonic polyps, we analyze local shape features (e.g. mean curvature, sphericity, etc.) of the colon’s surface for the generation of CADe candidates \cite{summers2005computed}. Even though \cite{summers2005computed} is a relatively straightforward approach for polyp detection compared to more recent data-driven colonic polyp CADe systems in the literature \cite{lu2011effective,Slabaugh2010}, it can serve as a sufficiently good candidate generation procedure when coupled with our random views of ConvNet observations and aggregation for effective false positive rejection.
%###################################################################################
\subsection{Cascaded CADe Architectures for False Positive Reduction} 
\label{sec:cascade}
There exist two types of cascaded CADe classification architectures for false positive reduction are two types: 1) extraction of new image features followed by retraining of a classifier on all candidates \cite{Yao2009,Slabaugh2010,seff20142d,roth2014new,lu2014computer} (from Sec. \ref{sec:candidates}) or 2) design of application dependent post-filtering components \cite{Barbu2006,lu2008,lu2009}. Different (often more computationally expensive) image features are calculated per extracted candidate, in order to reveal new information omitted from the CG step, since explicit brute-force search in CG is no longer necessary. Examples of heterogeneous CADe post-filters include the removal of 3D flexible tubes \cite{Barbu2006}, ileo-cecal valve \cite{lu2008} and extra-colonic findings \cite{lu2009} in CT colonography. Although training cascaded CADe systems using the same set of image features and the same type of classifier (e.g., SVM or random forest) is feasible, this approach  often demonstrates less effective overall performance (as discussed later) and is less employed. In this paper, we mainly exploit the first type of cascade, which uses deep ConvNet models as new components of integrated image feature representation and classification.
%###################################################################################
%###################################################################################
\section{Evaluation and Results}
\label{sec:results}
\subsection{Imaging Data Sets and Implementation}
\label{sec:image_data}
We evaluate our method on three medical imaging data sets that illustrate common clinical applications of CADe in CT imaging: sclerotic metastases in spine imaging, lymph nodes and colonic polyps in cancer monitoring and screening. We also show the scalability of ConvNets to different data set sizes, i.e. {\Npatientsbonelesions}, {\Npatientslymphnodes} (86 abdominal, 90 mediastinal) and {\Npatientspolyps} patients per data set respectively. Some statistics on patient population, total/mean (target) lesion numbers, total true positive (TP) and false positive (FP) candidate numbers, mean candidate numbers per case are given in Table \ref{tab:data_sets}. Note that one target can have several TP detections.
\begin{table*}[t]
  \centering
  \caption{CADe data sets used for evaluation: sclerotic metastases, lymph nodes, colonic polyps.}
    \begin{tabular}{lrrrrrr}
    \toprule
    \toprule
    \textbf{Dataset} & \textbf{\# Patients} & \textbf{\# Targets} & \textbf{\# TP} & \textbf{\# FP} & \textbf{\# Mean Targets} & \textbf{\# Mean Candidates} \\
    \midrule
    sclerotic lesions & {\Npatientsbonelesions} & 532 & 935   & 3,372   & 9.0 & 73.0\\
    lymph nodes       & {\Npatientslymphnodes}  & 983 & 1,966 & 6,692   & 5.6 & 49.2\\
    colonic polyps    & {\Npatientspolyps}      & 252 & 468   & 174,301 & 0.2 & 147.4\\
    \bottomrule 
		\bottomrule
    \end{tabular}%
  \label{tab:data_sets}%
\end{table*}%
%###################################################################################
For all imaging data sets used in this study, the image patches were centered at each CADe coordinate (of candidate VOI centroid from pre-existing CADe systems \cite{burns2013automated,liu2014mediastinal,cherry2014abdominal,summers2005computed}) with $32 \times 32$ pixels in resolution. All patches were sampled at 4 scales of $s = [30, 35, 40, 45]$ mm ROI edge length in physical image space, after isotropic resampling of the input CT images (see Fig. \ref{fig:roi_voi}). These scales cover the average dimensions for all objects of interest in the imaging data sets used in this study. Furthermore, all ROIs were randomly translated (up to 3 mm) and rotated at each scale (thus $N_s = 4$, $N_t = 5$ and $N_r = 5$), resulting in $N = 100$ image patches per ROI. Due to the much larger data set in the colonic polyp case, the parameters were chosen to be $N_s = 4$, $N_t = 2$ and $N_r = 5$), resulting in $N = 40$ image patches per ROI.

The training times for each ConvNet model were approximately 9-12 hours for the lymph node data set,  12-15 hours for the bone lesions data set, and 37 hours for the larger colonic polyps data set. All training was performed using a NVIDIA GeForce GTX TITAN (6GB on-board memory) for 1200 optimization epochs with unit Gaussian random parameter initializations as in  \cite{wan2013regularization}. Running $N=100$ 2D or 2.5D image patches at each ROI/VOI for classification of one CT volume only took circa 1-5 minutes. Image patch extraction from one CT volume lasted around 2 minutes at each scale. The employed ConvNet architecture is illustrated in Fig. \ref{fig:convnet}.
\begin{table}[H]
% \tiny
\footnotesize
  \centering
  \caption{Improvement with ConvNet Integration: previous$^1$ CADe performance compared to ConvNet$^2$ performance at the 3 FPs/patient rate.}
    \begin{tabular}{lrrrr}
    \toprule
    \toprule
    \textbf{Dataset} & \textbf{Sensitivity$^1$} & \textbf{Sensitivity$^2$} & \textbf{AUC$^1$} & \textbf{AUC$^2$} \\
    \midrule
    sclerotic lesions               & 57\%  & 70\%  & n/a   & 0.83 \\
    lymph nodes                     & 43\%  & 77\%  & 0.76  & 0.94 \\
    colonic polyps\tiny{($>=$6mm)}  & 58\%  & 75\%  & 0.79  & 0.82 \\
		colonic polyps\tiny{($>=$10mm)} & 92\%  & 98\%  & 0.94 & 0.99 \\
    \bottomrule
		\bottomrule
    \end{tabular}%
  \label{tab:sensitivities}%
\end{table}%
%###################################################################################
%###################################################################################
\subsection{Trained ConvNet Filter Kernels}
\begin{figure*}[t]
	\begin{center}
		\includegraphics[width=0.62\linewidth]{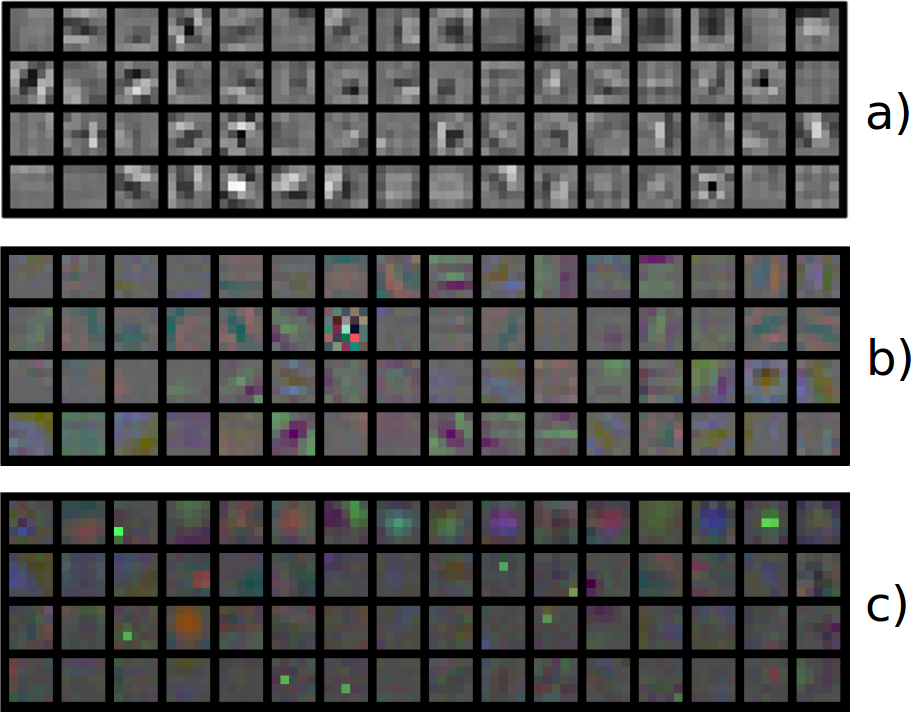}
		\end{center}
	\caption{The first layer of 64 learned convolutional kernels of a ConvNet trained on medical CT images on each of three different CT imaging data sets: a) sclerotic metastases, b) lymph nodes and c) colonic polyps. The color coding in b) and c) illustrates the filters kernels used in each orthogonal view when using our 2.5D approach. The learned convolutional filters for sclerotic metastases in a) are using one-channel as input only (encoded in gray scale and learned from axial CT images). Here, complex higher order gradients, blobness and difference of Gaussian filters dominate. In b,c), the convolutional filters for lymph nodes or colonic polyps are three-channels (encoded in RGB and trained using three orthogonal CT views per example). Kernels learned from lymph nodes are mostly blobness and gradients of different orientations/channels in b). Colonic polyp kernels in c) are visually more diversified than the filters in b), especially with new ``pointy'' patterns probably resembling polyp intrusions from 3D colonic surfaces or tips.}
	\label{fig:conv1}
\end{figure*}

The trained filters of the first convolutional layer for all three imaging data sets used in this study can be seen in Fig. \ref{fig:conv1}. A mixed set of low and high frequency patterns exists in the first convolutional layer. The filter kernels ``capture'' the essential information that is necessary for each classification task. These automatically learned filters need no tuning by hand, and thus have a major advantage over more traditional CADe approaches \cite{jones2014computer}. In Fig. \ref{fig:conv1} a), the learned convolutional filters for sclerotic metastases are one-channel only (encoded in gray scale and learned from axial CT images); b,c), the convolutional filters for lymph nodes or colonic polyps are three-channels (encoded in RGB and trained using three orthogonal CT views per example). Different visual characteristics of ConvNet filter kernels are discussed in Fig. \ref{fig:conv1} as well.   
%###################################################################################
%###################################################################################
%###################################################################################
\subsection{2D, 2.5D and 3D ConvNet Configurations}
In this experiment, we compare the CADe performance of varying dimensional inputs to that of our ConvNet architecture: 2D ROIs, the proposed 2.5D VOIs and 3D VOI stacks. The effect of data augmentation for ConvNet training is evaluated on the abdominal lymph node data set. An 80\%/20\% split of 86 patients is used for training and testing, respectively. Fig. \ref{fig:lymphnodes_2D_2-5D_3D} shows the FROC performance for both training (Left) and testing (Right). It can be observed that a pure 2.5D approach on the original CT data is not sufficient to capture the variety of lymph nodes in the test set. However, adding the proposed random observations in both training and testing (as a form of data augmentation) leads to the best performing CADe framework at a level of 3 FPs/vol., compared to 2D and 3D approaches. 

In the 3D case, we extract full $32\times 32\times 32$ VOI image stacks as input to our ConvNet. In this case, the amount of training data is also not enough to learn all parameters of the ConvNet without data augmentation in order to generalize well to the testing data. Clear overfitting occurs in testing, highlighting the advantages of using a 2.5D approach in applications where training data can be too limited (as in many medical imaging problems). Yet, adding data augmentation to the training set improves the performance in 3D markedly with the trade-off of adding $\sim 4\times$ more training time in order to achieve convergence (see Table \ref{tab:training_times}), and performs only comparable to the augmented 2.5D case.
%###################################################################################
\begin{table}[H]
\scriptsize
  \centering
  \caption{Training times until convergence in the 2D vs. 2.5D vs. 3D cases on the abdominal lymph node data set:}
    \begin{tabular}{lrr}
    \toprule
    \toprule
    \textbf{Input Dimensions} & \textbf{Augmentation} & \textbf{Time (min)}\\
    \midrule
   2D & no  & 123\\
   2D & yes & 847\\
   2.5D & no & 59\\
   2.5D & yes & 476\\
   3D & no & 119\\
   3D & yes &  1844\\
    \bottomrule 
		\bottomrule
    \end{tabular}%
  \label{tab:training_times}%
\end{table}%

%###################################################################################
\subsection{Detection of Sclerotic Metastases}
\label{sec:results_bonelesions}
In our evaluation, radiologists labeled a total of 532 sclerotic metastases in CT images of 49 patients (14 female, 35 male patients; mean age 57.0 years; age range of 12-77 years). A lesion is only labeled if its volume is greater than 300 mm$^3$. These CT scans have reconstruction slice thicknesses ranging between 2.5 mm and 5 mm. Furthermore, we include 10 control cases (4 female, 6 male patients; mean age 55.2 years; age range of 19-70 years) without any spinal lesions. Note that 2.5-5 mm thick-sliced CT volumes are used for this study (for low dose CT radiation). Due to this relatively large slice thickness, our spatial transformations are all drawn from within the axial plane, i.e. following the 2D approach introduced in Sec. \ref{sec:roi_voi}. Coronal or Sagittal image views demonstrate low longitudinal resolutions and thus have poor diagnostic quality.

Any false-positive detection from the candidate generation step on these patients is used as a ``negative'' candidate example in training the ConvNet. This strategy would be considered as ``hard negative mining'' or ``bootstrapping'' in the general computer vision or statistics literature. The maximum sensitivity of this candidate generation step in testing was 88.9\% \cite{burns2013automated}. All patients were randomly split into five sets at the patient level in order to allow a 5-fold cross-validation. We adjust the sample rates for positive and negative image patches in order to generate a balanced data set for training (i.e., 50\% positives and 50\% negatives). This means all randomly sampled positives are included in training, but only a subset of negative random samples are used. Balancing between positive and negative training populations is generally beneficial for training ConvNets when optimizing with logistic regression cost \cite{lecun1989backpropagation,krizhevsky2012imagenet}. For this data set, a 2D approach is used: each 2D image patch was centered at the CADe coordinate with $32 \times 32$ pixels in resolution. As stated in Sec. \ref{sec:image_data}, all patches are sampled at 4 scales of $s = [30, 35, 40, 45]$ mm ROI edge length in the physical image space, after isotropic resampling of the CT images (see Fig. \ref{fig:roi_voi}). In this data set, we use a bone window level of [-250, 1250 HU]. 
%###################################################################################
\begin{figure}[t]
	\centering
		\includegraphics[width=8.5cm]{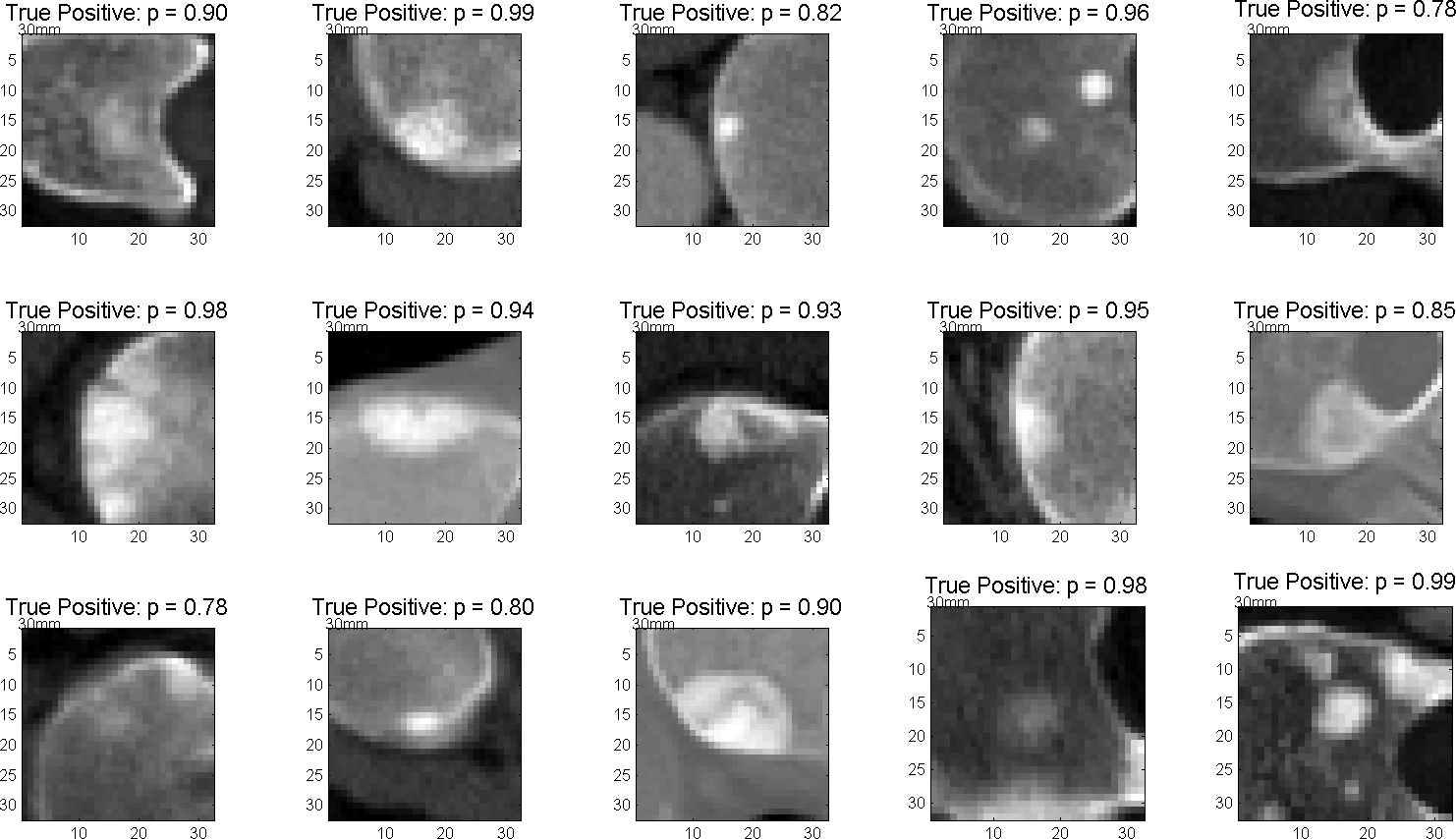}
	\caption{Detection of sclerotic metastases: test probabilities of the ConvNet for being sclerotic metastases on `true' sclerotic metastases candidate examples (1.0 equals 100\% probability of representing a true positive).}
	\label{fig:lymphnode_true_ConvNet_predictions}
\end{figure}
%###################################################################################
\begin{figure}[t]
	\centering
		\includegraphics[width=8.5cm]{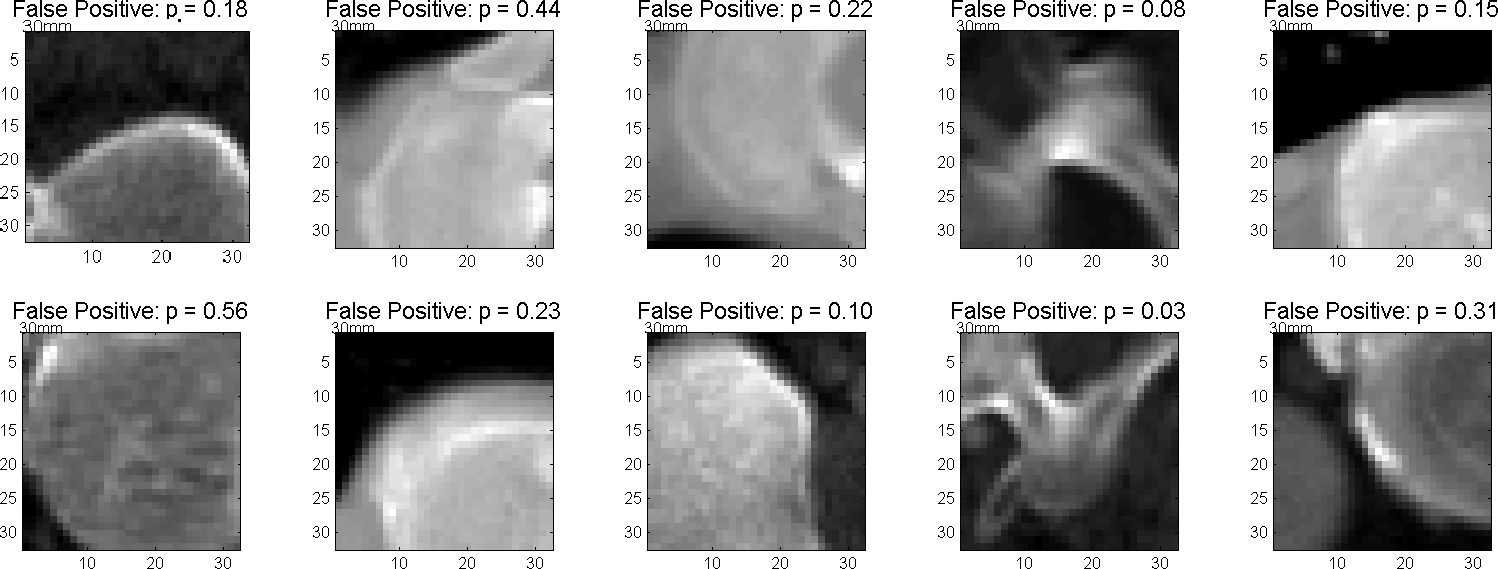}
	\caption{Detection of sclerotic metastases: test probabilities of the ConvNet for being sclerotic metastases on `false' sclerotic metastases candidate examples (0.0 equals 100\% probability of representing a false positive).}
	\label{fig:lymphnode_false_ConvNet_predictions}
\end{figure}
%###################################################################################
We now apply the trained ConvNet to classify image patches from the test data sets. Figure \ref{fig:lymphnode_true_ConvNet_predictions} and Fig. \ref{fig:lymphnode_false_ConvNet_predictions} show typical classification probabilities on two random subsets of positive and negative ROIs in the test case, respectively.

\begin{figure}[t]
	\centering
		\includegraphics[width=8.5cm]{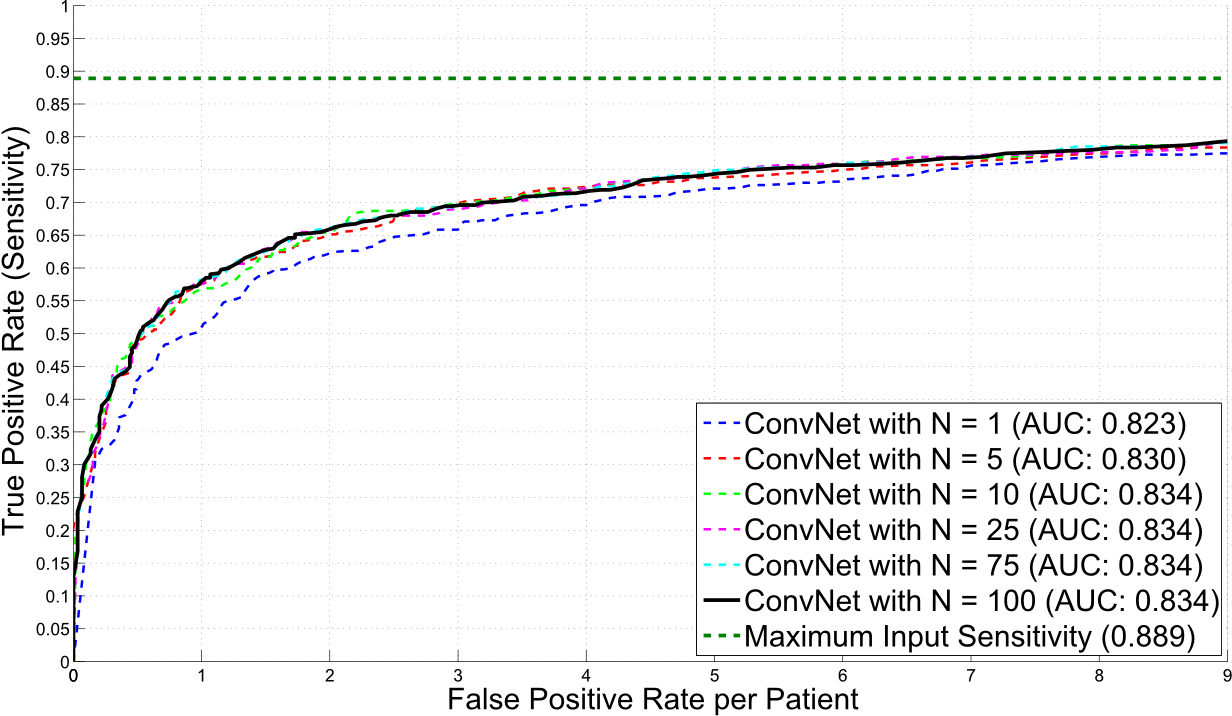}
	\caption{Detection of sclerotic metastases: FROC curves for a 5-fold cross-validation using varying numbers of $N$ random view ConvNet observations in testing of 59 patients (49 with sclerotic metastases and 10 normal controls). AUC values are computed for corresponding ROC curves.}
	\label{fig:froc_varying_N}
\end{figure}

Averaging the $N$ predictions at each CADe candidate allows us to compute a per-candidate probability $p(x)$, as in Eq. \ref{equ:prob}. Varying thresholds on probability $p(x)$ are used to compute Free-Response Receiver Operating Characteristic (FROC) curves. FROC curves are compared in Fig. \ref{fig:froc_varying_N} for the configurations of varying $N$ and demonstrate that the classification performance saturates quickly with increasing $N$. If $N<100$, we use a random subset of observations to compute the average prediction value. This means the run-time efficiency of our second layer detection could be further improved without losing noticeable performance by decreasing $N$. The proposed method reduces the number of FPs/patient of the existing sclerotic metastases CADe systems \cite{burns2013automated} from 4 to 1.2, 7 to 3, and 12 to 9.5 when comparing sensitivity rates of 60\%, 70\%, and 80\% respectively in cross-validation testing (at $N = 100$). The Area-Under-the-Curve (AUC) values remain stable at 0.834 for $N$ between $\left[10,...,100\right]$. 

\begin{figure}[t]
	\centering
		\includegraphics[width=8.5cm]{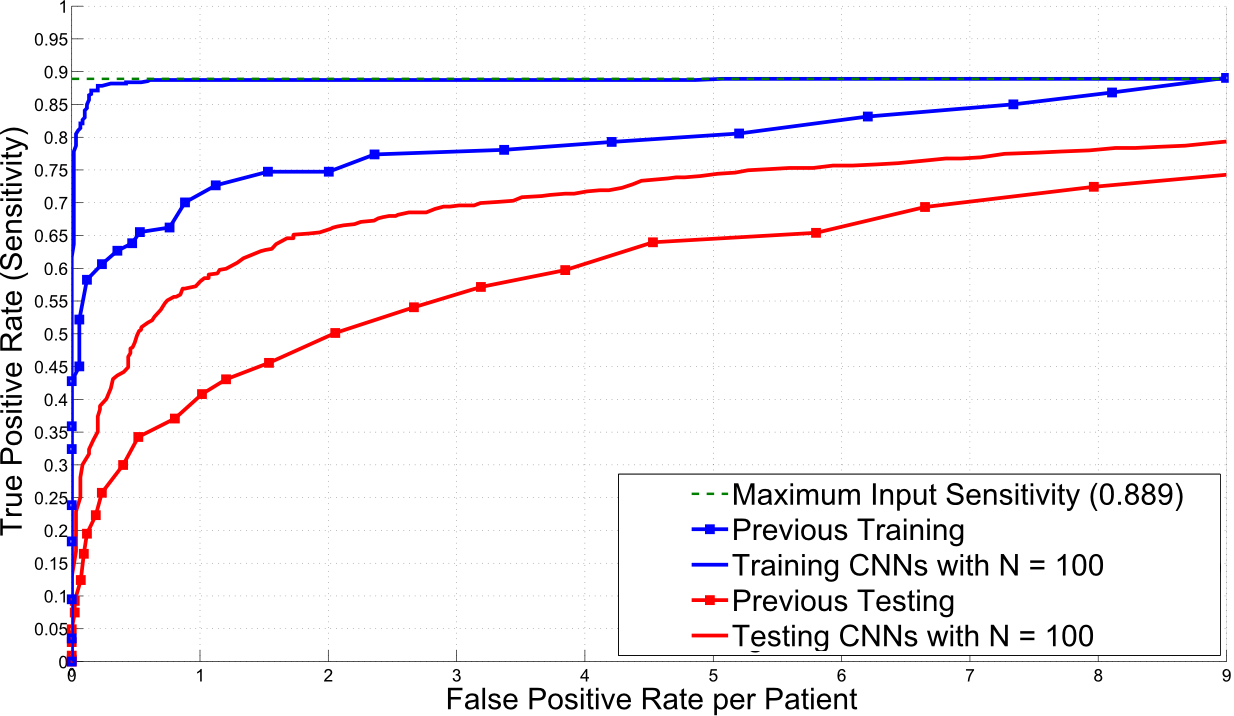}
	\caption{Detection of sclerotic metastases: comparison of FROC curves of the initial bone lesion candidate generation (squares) compared to the final classification using $N$ = 100 random view ConvNet observations (lines) for both training and testing cases. Results are computed using a 5-fold cross-validation in 59 patients (49 with sclerotic metastases and 10 normal controls).}
	\label{fig:froc_comparison}
\end{figure}

Fig. \ref{fig:froc_comparison} compares the FROCs from the initial (first layer) CADe system \cite{burns2013automated} and illustrates the progression towards the proposed coarse-to-fine two tiered method in both training and testing datasets. This clearly demonstrates a marked improvement in performance. The FROC performance differences from training to testing in both cases still show some degree of overfitting, which can be addressed by including more patient data (59 patients are in general too few to train ConvNets to generalize well). This observation is insightful for later work on deep learning system design for medical diagnosis. 
%###################################################################################
%###################################################################################
%###################################################################################
%###################################################################################
\subsection{Detection of Thoracoabdominal Lymph Nodes}
\label{sec:results_lymphnodes}
\begin{figure}[t]
	\centering
		\includegraphics[width=8.5cm]{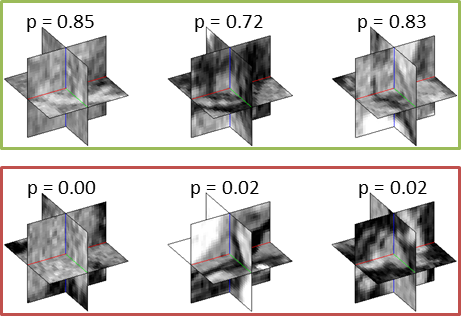}
	\caption{Detection of lymph nodes: test probabilities of the ConvNet for being a lymph node on `true' (top box) and `false' (bottom box) lymph node candidate examples.}
	\label{fig:lymphnode_ConvNet_predictions}
\end{figure}

\begin{figure}[t]
	\centering
		\includegraphics[width=8.5cm]{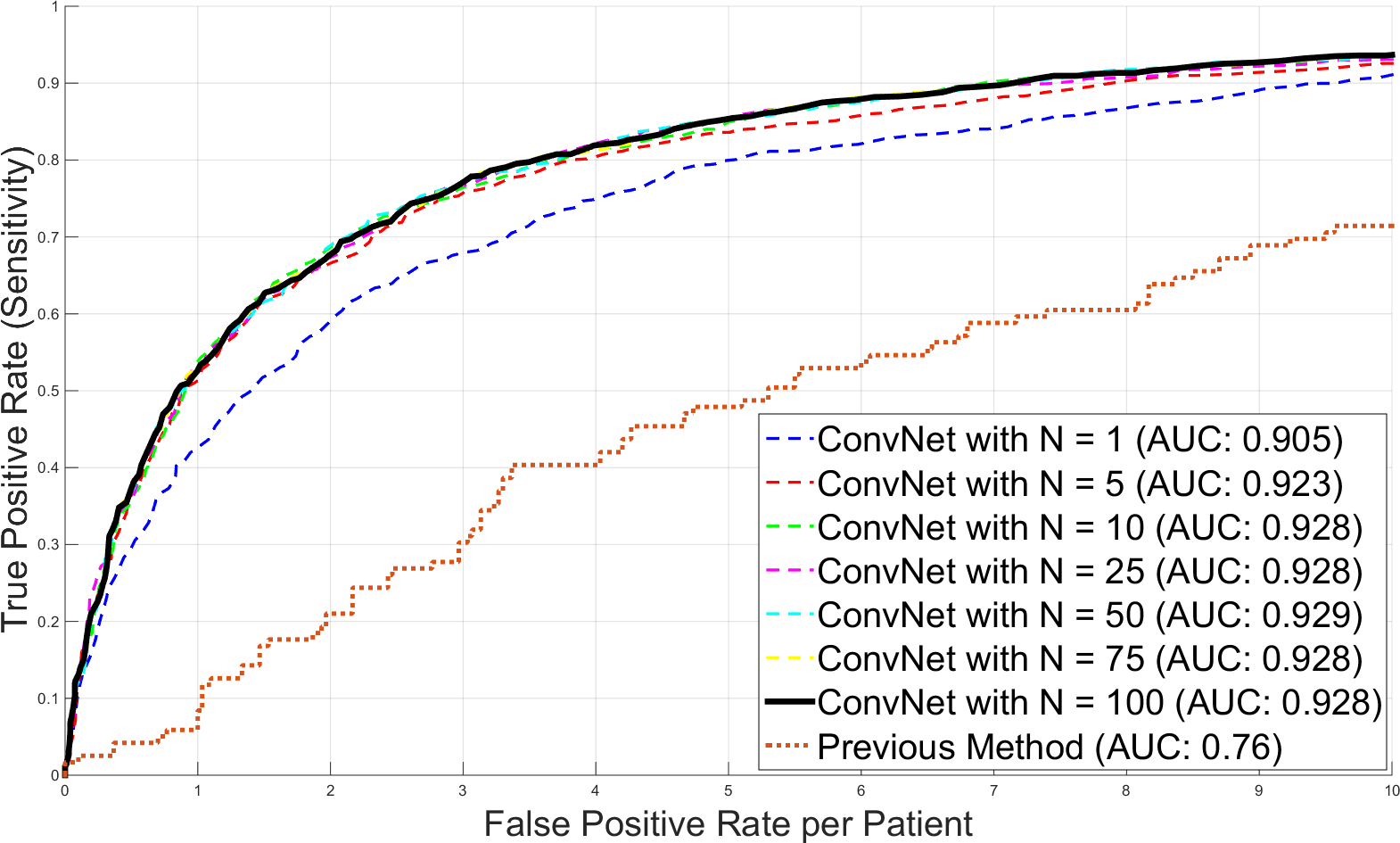}
	\caption{Detection of lymph nodes: FROC curves for a 3-fold cross-validation using a varying number of $N$ random view ConvNet observations in 176 patients. AUC values are computed for corresponding ROC curves. The previous performance by \cite{cherry2014abdominal} is shown for comparison.}
	\label{fig:roc_froc}
\end{figure}

The next data set consists of 176 patients that are used for CADe of lymph nodes. Here, the slice thickness of CT scans was $\leq$1 mm. Hence, we were able to apply a 2.5D approach (composite of three orthogonal 2D views) for sampling each CADe candidate as described in Sec. \ref{sec:roi_voi}. Radiologists labeled a total of 388 mediastinal lymph nodes and 595 abdominal lymph nodes as `positives' in the CT images. In order to objectively evaluate the performance of our ConvNet based 2.5D detection approach, 100\% sensitivity at the lymph node candidate generation stage for training is assumed by injecting the labeled lymph nodes into the set of CADe lymph node candidates (see Sec. \ref{sec:candidates}). The CADe system produces a total of 6,692 false-positive detections ($>$15 mm away from true lymph node) in the mediastinum and the abdomen. These false-positive detections are used as `negative' lymph node candidate examples for training the ConvNets. There are a total of 1956 true-positive detections from \cite{liu2014mediastinal,cherry2014abdominal}. All patients are randomly split into three subsets (at the patient level) to allow a 3-fold cross-validation. We use different sample rates of positive and negative image patches to generate a balanced training set. This proves beneficial for training the ConvNet. Each three-channel image patch (as a 2.5D view) is centered at a CADe coordinate with $32 \times 32$ pixels. Again, all patches are sampled at 4 scales: $s=\left[30,35,40,45\right]$ mm for the VOI edge length in the physical image space, after isotropic resampling of the CT images (see Fig. \ref{fig:roi_voi}). We use a soft-tissue window level of [-100, 200 HU] as in \cite{barbu2012automatic}. Furthermore, all VOIs are $N=100$ times randomly translated (up to 3 mm) and rotated at each scale. After training, we apply the trained ConvNet to classify image patches from the testing datasets. Figure \ref{fig:lymphnode_ConvNet_predictions} shows some typical classification probabilities on a random subset of test VOIs.
%###################################################################################
%###################################################################################
Averaging the $N$ predictions at each lymph node candidate allows us to compute a per-candidate probability $p(x)$, as in Eq. \ref{equ:prob}. Varying a threshold parameter on this probability allows us to compute the free-response receiver operating characteristic (FROC) curves. Different FROC curves are compared in Fig. \ref{fig:roc_froc} with varying $N$. It can be observed that the classification performance saturates quickly with increasing $N$, consistent with Sec. \ref{sec:results_bonelesions}. The classification sensitivity improves on the existing lymph node CADe systems \cite{liu2014mediastinal,cherry2014abdominal} from 55\% to 70\% in the mediastinum and from 30\% to 83\% in the abdomen at a low rate of 3 FP per patient volume (FP/vol.), for $N=100$ \cite{roth2014new}. The AUC improves from 0.76 to 0.942 in the abdomen, when using the proposed false-positive reduction approach (AUC for the mediastinal lymph nodes was not available for comparison). At an operating point of 3 FP/vol., we achieve significant improvement: $p<0.001$ in both mediastinum and abdomen, respectively (Fisher's exact test). 

Further experiments show that performing a joint ConvNet model trained on both mediastinal and abdominal lymph node candidates together can improve the classification by $\sim$10\% to $\sim$80\% sensitivity improvements (case by case) at 3 FP/vol. in the mediastinal set. The overall 70\% sensitivity at 3 FP/vol. increases to 77\% in the mediastinum. The sensitivity level in the abdomen datasets remains stable.
%###################################################################################
%###################################################################################
We achieve a substantial improvement compared to the state-of-the-art methods in lymph node detection. \cite{feulner2013lymph} reports a 52.9\% sensitivity rate at 3.1 FP/vol. in the mediastinum, while achieving a rate of 70\% \cite{roth2014new} or 77\% (joint training) at 3 FP/vol. In the abdomen, the most recent work (\cite{nakamura2013automatic}) shows a 70.5\% sensitivity rate at 13.0 FP/vol. We obtain 83\% at 3 FP/vol. (assuming $\sim$100\% sensitivity at the lymph node candidate generation stage). Note that any direct comparison to another recent work is difficult since common datasets were not previously utilized. Therefore, our data set\footnote{\url{http://www.cc.nih.gov/about/SeniorStaff/ronald_summers.html}}\footnote{\url{http://dx.doi.org/10.7937/K9/TCIA.2015.AQIIDCNM}} and supporting material\footnote{\url{www.holgerroth.com}} have been made publicly available for future comparison purposes. 
%###################################################################################
\subsection{2.5D ConvNets Compared to Shallow Classification}
\begin{figure}[H]
	\centering
		\includegraphics[width=8.5cm]{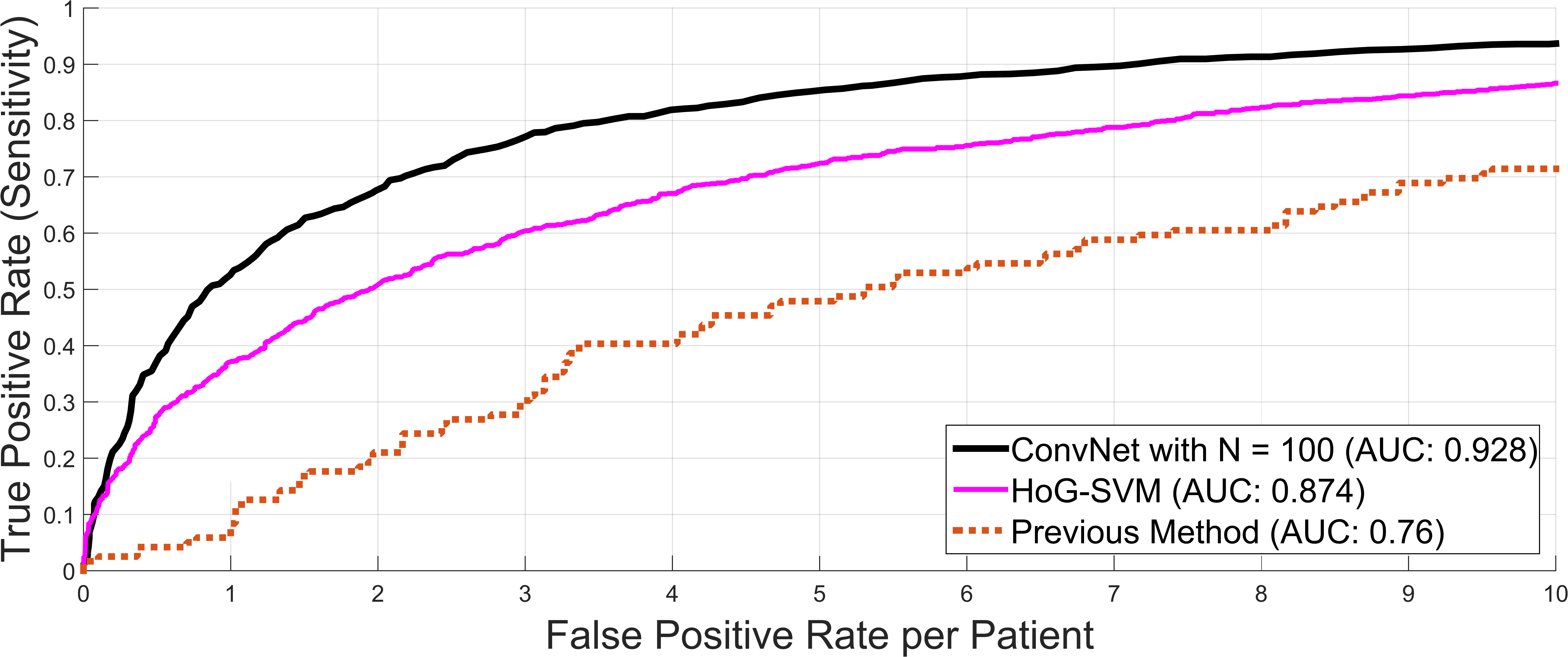}
	\caption{Comparison of the FROC performance of the previous method as candidate generation step using a random forest classifier \cite{cherry2014abdominal} against an alternative second level classification approach using histogram of oriented gradients (HoG) \cite{seff20142d} and the proposed 2.5D ConvNet approach using ConvNet observation on $N=100$ random views.}
	\label{fig:lymphnodes_cnn_vs_hog_froc}
\end{figure}

We compare our 2.5D approach to other means of second tier classification (FP filter or ``killer''), e.g., linear SVM based on Histogram of Oriented Gradients (HoG) features as proposed in \cite{seff20142d}. Here, both simple pooling and sparse linear decision fusion schemes to aggregate 2D detection scores are exploited for the final 3D lymph node detection. This type of cascade classification is similar in spirit to our presented second tier deep classifier (ConvNet), but uses state-of-the-art shallow classifiers (libSVM \cite{Chang2011} and sparse linear fusion via the Relevance Vector Machine \cite{Raykar2008}).  As shown in Fig. \ref{fig:lymphnodes_cnn_vs_hog_froc}, a clear advantage of using the proposed 2.5D ConvNet method can be observed (unlike in \cite{seff20142d}). Note that this shallow linear cascade approach via new image features, such as Histogram of Oriented Gradients, already significantly surpasses previous state-of-the-art methods \cite{feulner2013lymph,nakamura2013automatic,feuerstein2009automatic}. Furthermore, we use the same set of image features and random forest classifiers in a two-tiered cascade of hierarchy \cite{cherry2014abdominal}. No improvement in CADe performance is observed. This highlights the importance of leveraging heterogeneous image features in the two stages of candidate generation and candidate classification.
%###################################################################################
%###################################################################################
\subsection{3D, 2D or 2.5D ConvNets: Alleviating Curse-of-dimensionality via Random View Aggregation}
\begin{figure*}[t]
	\centering
		\includegraphics[width=17cm]{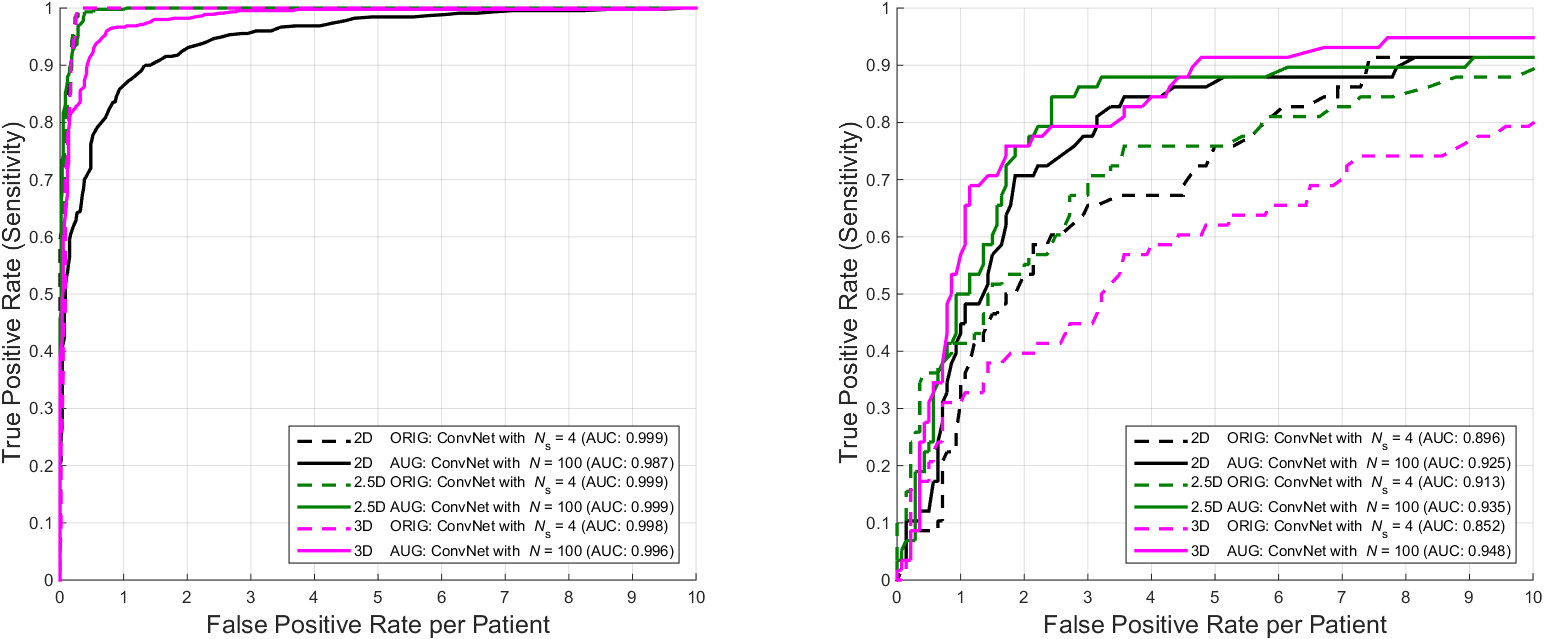}
	\caption{Comparison of the FROC performance of when training a ConvNet with 2D, 2.5D and 3D inputs of the original (``ORIG'') or augmented (``AUG'') CT data. In the ``ORIG'' setting, 2D ConvNet shows the best generalized testing FROC result, followed by 3D and 2.5D ConvNets. The 2.5D approach using aggregation of random observations (``AUG'') in both training (Left) and testing (Right), out-performs both 2D and 3D approaches on the original data at the 3 FPs/patient level. The 2.5D ConvNet trained on augmented data overall performs comparably to a more computationally expensive 3D ConvNet approach on augmented 3D inputs. In brief, the evaluated 2.5D ``AUG'' ConvNet is chosen as the best trade-off lymph node detection model between effectiveness and efficiency.}
	\label{fig:lymphnodes_2D_2-5D_3D}
\end{figure*}

Medical images are intrinsically 3D, but relative to other computer vision problems, CADe problems often lack sufficient training data to learn 3D models effectively (see Fig. \ref{fig:lymphnodes_2D_2-5D_3D}). From the perspective of the `curse of dimensionality', a 3D task requires at least one order of magnitude more training data than a 2D task. This problematic data distribution setting can hamper the performance of learning algorithms in CADe, thus motivating us to exploit the 2D/2.5D decompositional sampling and aggregation representation. The number of training instances has been increased up to 100 times (although not independent and identically distributed samples) for training ConvNets, without directly learning the complex and explicit 3D object representation and classification. Likewise, the compositional two-stream 2D ConvNet models run on separate spatial (RGB) and temporal (i.e., optical flow field) video frames and achieve the mean accuracy of 87.9\% in action classification task, based on a middle scale dataset UCF-101 \cite{Simonyan2014Two}. This result significantly outperforms the direct 3D ``spatial-temporal'' ConvNet method \cite{Karpathy2014} at 65.4\% (mean accuracy), evaluated on the same UCF-101 benchmark. 

In Fig. \ref{fig:lymphnodes_2D_2-5D_3D}, we conduct extensive empirical evaluation and comparative study using 3D, 2D or 2.5D ConvNets for lymph node detection. 1), The ``ORIG'' versions of 3D, 2D or 2.5D ConvNets demonstrate consistently better training performance than the ``AUG'' setting (i.e., more data in ``AUG'' cause harder to over-fit), as illustrated in Fig. \ref{fig:lymphnodes_2D_2-5D_3D} {\bf Left}. However in testing, 3D, 2D or 2.5D ConvNets trained under data augmentation or ``AUG'' all clearly outperform their ``ORIG'' counterparts. 2). Without data augmentation, the more complex 3D ConvNet model shows a great decline in performance between training to testing compared to the 2D and 2.5D ConvNets, which indicates stronger over-fitting due to curse-of-dimensionality (Fig. \ref{fig:lymphnodes_2D_2-5D_3D} {\bf Right}). In the ``ORIG'' setting, 2.5D and 2D ConvNets give noticeably better testing FROC results (while being comparable overall between themselves), followed by the 3D ConvNet. Consequently, this observation validates the concept that simpler or lower-dimensional learning models generalize better than complex ones without sufficient available training data (as in ``ORIG'' setting). 3). Data augmentation based on random view aggregation, as proposed in our original work (\cite{roth2014new}), effectively circumvents the ``curse-of-dimensionality'' or ``over-fitting'' issue in the data-demanding ConvNet training procedures. This strategy has been adapted to computer-aided pulmonary embolism detection (\cite{Tajbakhsh2015}), lung nodule classification (\cite{Ginneken2015,Shen2015}) in CT images and polyp detection in colonoscopy videos (\cite{Park2015,Tajbakhsh2015b}). 4), The 2.5D and 3D (``AUG'') ConvNets dominate 2D (``AUG'') ConvNet in most of FROC ranges; while 2.5D ConvNet performs the best in the FP range of [2-4] than the other two models. Overall 2.5D ConvNet performs comparably (in both training and testing) to the more computationally expensive 3D ConvNet configuration, as augmented 3D volumetric VOI inputs are required. In summary, the evaluated 2.5D ``AUG'' ConvNet is selected as the best trade-off lymph node detection model, when detection performance and computational efficiency are taken into account.
%###################################################################################
%###################################################################################
%###################################################################################
%###################################################################################
\subsection{Detection of Colonic Polyps}
\label{sec:resutls_polyps}
In CT colonography (CTC), patients are typically scanned in the prone and supine positions \cite{johnson2008accuracy}, so we obtain two CT volumes per patient study. We use CTC images from three institutions in this study. A total of {\Npatientspolyps} patients with prone and supine CTC images were included (as in \cite{summers2005computed}). In this data set, each polyp $\geq$6 mm found at optical colonoscopy was located on the prone and supine CTC examinations using 3D endoluminal colon renderings with ``fly-through'' viewing and multiplanar reformatted images. 

The patients were separated into training ($n={\Npatientspolypstraining}$) and testing sets ($n={\Npatientspolypstesting}$) with similar age and gender distributions -- an approximate 1:2 split. There were 79 training and 173 testing polyps ($>=$6mm); and 22 training and 37 testing polyps ($>=$10mm, considered as large polyps) in our CTC dataset \cite{summers2005computed}. The candidate generation step for colonic polyps is performed by the CADe system presented in \cite{summers2005computed}. In this system, the colonic wall and lumen are first segmented, and any tagged colonic fluids were removed. To identify colonic polyps, the 3D colon surface undergoes an examination on shape filtering features to generate CADe findings or candidates \cite{summers2005computed}. 

\begin{figure}[t]
	\centering
		\includegraphics[width=8.5cm]{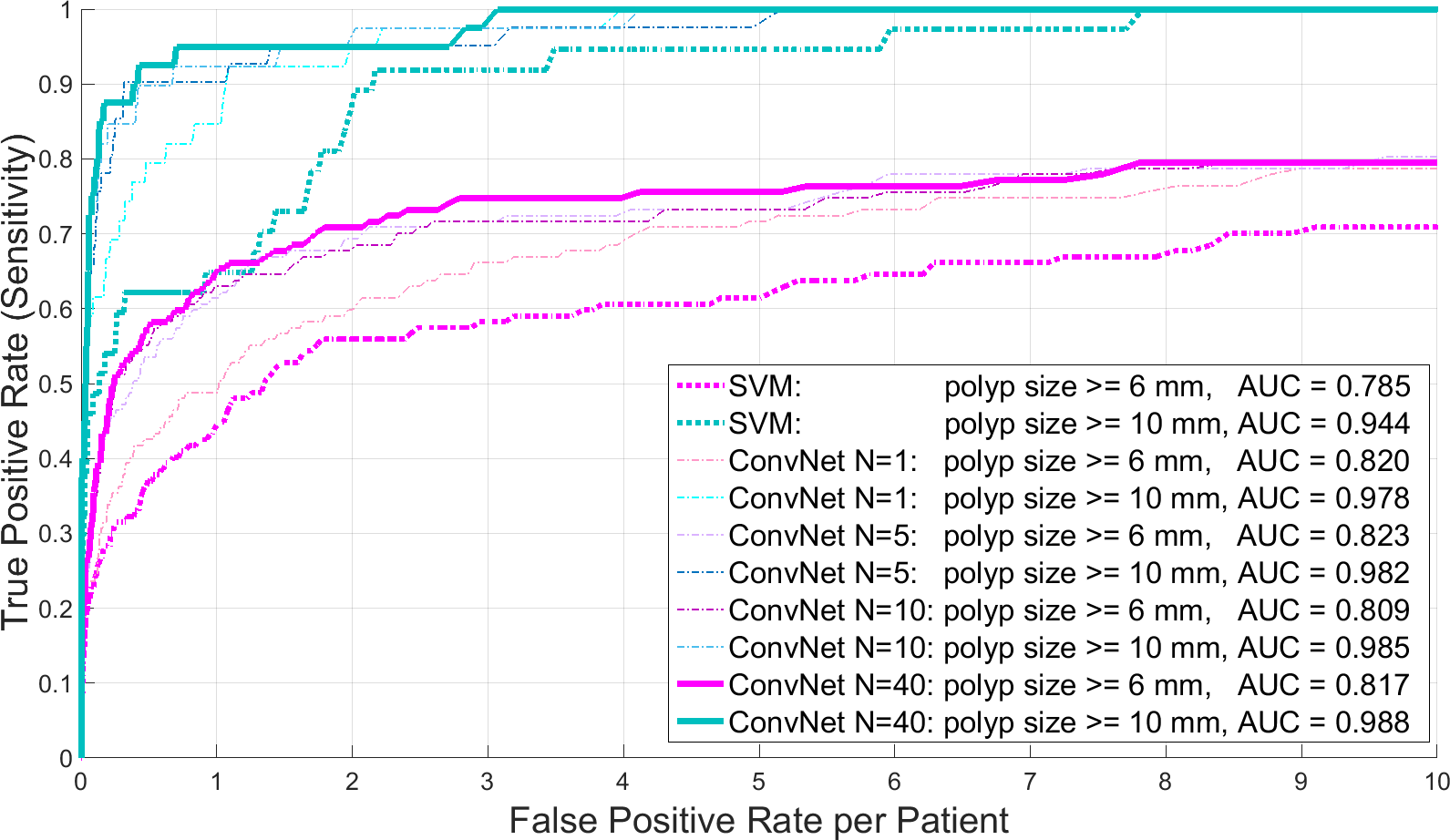}
	\caption{Detection of colonic polyps: FROC curves for different polyp sizes, using up to $N=40$ random view ConvNet observations in {\Npatientspolypstesting} testing CT colonography patients.}
	\label{fig:polyp_froc}
\end{figure}

The FROC curves for detecting adenomatous polyps of $\geq$6 and $\geq$10 mm, respectively, are shown in Fig. \ref{fig:polyp_froc} for a varying number of observations $N$. The performance saturates quickly after $N=10$ random observations. At both polyp size thresholds, a large improvement in sensitivity at all false-positive rates can be observed. In all cases, the sensitivity levels were higher for larger polyps at constant false-positive rates. At a rate of 3 FPs per patient for polyps $\geq$6 mm, the sensitivities per patient were raised from 58\% using a SVM classifier (as in \cite{summers2005computed}) to 75\% using our 2.5D ConvNet approach (see Table \ref{tab:sensitivities}). These results are comparable to other already highly tuned CADe systems for colonic polyp detection in CTC, such as \cite{lu2011effective,lu2014computer,Slabaugh2010}. 

Note that our system achieves significantly higher sensitivities of 95\%, 98\% at 1 or 3 FP/vol. for {\em clinically actionable $\geq$10 mm polyps}, compared to sensitivities of 82\% at 3.65 FP/vol. in \cite{lu2011effective} and 76\% at 1 FP/vol.; 95\% at 4.5 FP/vol. for \cite{Slabaugh2010}. The hierarchical voxel labeling CADe approaches for colonic polyps \cite{lu2011effective,lu2014computer} better handle smaller polyps ($\geq$6 but $<10$ mm), at 84.7\% sensitivity with less than 3.62 FP/vol. but exhibit inferior performance on clinically more important and relevant large polyps. Note that the results between our work and previous methods \cite{lu2011effective,lu2014computer,Slabaugh2010} are not possible to be strictly compared since different datasets are evaluated. The colonic polyp CADe dataset scales are similar: 770 tagged-prep CT scans from multiple medical sites (358 training and 412 validation) in \cite{lu2011effective,lu2014computer}; 180 patients (360 CTC volumes) for training and 202 patients (404 volumes) for testing \cite{Slabaugh2010}. 

Finally, operating at 1 FP/patient to obtain about 95\% sensitivity in testing (improved from $\sim$65\% in \cite{summers2005computed}) for $\geq$10 mm large polyp detection is a desirable clinical setting for employing CADe as a second reader mode, with a minimal extra burden for radiologists. In \cite{Slabaugh2010}, approximately four times more effort is needed to review FPs (i.e., retaining 95\% sensitivity at 4.5 FP/vol.).
%###################################################################################
%###################################################################################
%###################################################################################
\subsection{Limitation \& Improvement} Although consistent FROC improvements are observed in Fig. \ref{fig:polyp_froc} for both polyp categories of $\geq$6 and $\geq$10 mm, our final system demonstrates more appealing performance for large polyps (i.e., $\geq$10 mm). Achieving 95\% sensitivity at 1 FP/patient. in testing is the best reported quantitative benchmark, to the best of our knowledge, for a large-scale colonic polyp CADe system. For polyps between 6 and 9 mm, our random 2.5D view sampling may not be optimal due to the smaller object size to detect (a portion of sampled 2.5 images may contain only some tiny fields-of-view of the target polyp). Potentially, the performance could be improved by adopting a local colonic surface alignment, such as \cite{Yao2009}, to further guide and constrain our random view sampling procedure.
%###################################################################################
%###################################################################################
\section{Discussion and Conclusions}
This work (among others, such as \cite{prasoon2013deep} and \cite{cirecsan2013mitosis}) reveals that deep ConvNets can be extended to 2D and 3D medical image analysis tasks. We demonstrate significant improvements on CADe performance of three pathology categories (i.e., bone lesions, enlarged lymph nodes and colonic polyps) using CT images. Building upon existing CADe systems, we show that a random set of ConvNet observations (via both 2D and 2.5D approaches) can be exploited to drastically improve the sensitivities over various false-positive rates from initial CADe detections. Sampling at different scales, random translations and rotations around each of the CADe detections can be employed to prevent or alleviate overfitting during training and increase the ConvNet's classification performance. Subsequently, the testing FROC curves exhibit marked improvements on sensitivity levels at the range of clinically relevant FP/vol. rates in all three evaluated CT imaging data sets. Furthermore, our results indicate that ConvNets can improve the state-of-the-art (as in the case of lymph nodes) or are at least comparable to already highly tuned CADe systems, as in the case of colonic polyp detection \cite{lu2011effective,lu2014computer,Slabaugh2010}. 

The main purpose of a 2.5D approach is to decompose the volumetric information from each VOI into a set of random 2.5D images (with three channels) that combine the orthogonal slices at $N$ reformatted orientations, in the original 3D imaging space. Our relatively simple re-sampling of the 3D data circumvents the usage of 3D ConvNets directly \cite{turaga2010convolutional}. This not only greatly reduces the computational burden for training and testing, but also more importantly, alleviates the curse-of-dimensionality problem. Direct training of 3D deep ConvNets \cite{turaga2010convolutional} for a volumetric object detection problem may currently cause scalability issues when data augmentation is not feasible or often severe lack of sufficient training samples, especially in the medical imaging domain. ConvNets generally need tremendous amounts of training examples to address the overfitting issue, with respect to the large number of model parameters. Data augmentation can be useful, as shown in this study, but the trade-off between computational burden and classification needs to be made. A 2.5D approach as proposed here can be a valid alternative to using 3D inputs. Random resampling is an effective and efficient way to increase the amount of available training data  in 3D, as in the presented approach. \cite{krizhevsky2012imagenet} uses translational shifting and mirroring of 2D image patches for this purpose. Our 2.5D representation is intuitive and applies the success of large-scale 2D image classification, using ConvNets \cite{krizhevsky2012imagenet} effortlessly into 3D space. The above averaging process (i.e., Eq. \ref{equ:prob}) further improves the robustness and stability of 2D/2.5D ConvNet labeling on random views in validation or testing (see Sec. \ref{sec:results}).

A secondary advantage of using 2.5D inputs may be that ConvNets that are pre-trained on larger data bases available in the computer vision domain (such as \textit{ImageNet}) could be used. Potentially allowing the ConvNet optimization to start from an initialization that is better than starting from Gaussian random parameters \cite{yosinski2014transferable,girshick2014rich}.

Potentially, larger and deeper convolutional neural networks could be applied to further improve classification performance \cite{simonyan2014very,Szegedy2014Going}. However, the curse-of-dimensionality problem makes it difficult to assess the amount of necessary data that is needed to effectively train these very deep networks. Extensions of ConvNets to 3D have been proposed, but computational cost and memory consumption can be still too high to efficiently implement them on current computer graphics hardware units \cite{turaga2010convolutional}. 

Finally, the proposed 2D and 2.5D generalization of ConvNets is promising for various applications in computer-aided detection of 3D medical images. For example, the 2D views with the highest probability of containing a lesion could be used to present ``classifier-guided'' reformatted visualizations at that orientation (optimal to the ConvNet) to assist in radiologists' reading. In summary, we present and validate the use of 3D VOIs with a new 2D and 2.5D representation that may easily facilitate a generally purposed 3D object detection-by-classification scheme.
%###################################################################################
\vspace{5 mm}
\section*{Acknowledgment}
This work was supported by the Intramural Research Program of the NIH Clinical Center. We would like to thank Ms. Isabella Nogues for proofreading this article.
%###################################################################################
%\clearpage
%\newpage
%\vspace{10 mm}
\bibliographystyle{ieeetr} % (1) prints author names abbreviated (like {abbrv}) in the references section and (2) sorts references numerically in citation order (like {unsrt})
\bibliography{references_tmi}

\vfill

\end{document}